\documentclass{article}
\usepackage{spconf,amsmath,graphicx}
\usepackage{booktabs}

\title{A Fine-tuned Wav2vec 2.0/HuBERT Benchmark For Speech Emotion Recognition, Speaker Verification and Spoken Language Understanding}
\name{Yingzhi Wang$^{\star}$ \qquad Abdelmoumene Boumadane$^{\star}$ \qquad Abdelwahab Heba$^{\star}$}
\address{$^{\star}$ Zaion Lab, Zaion, Paris, France}

\begin{document}
%
\maketitle
\begin{abstract}
Speech self-supervised models such as wav2vec 2.0 and HuBERT are making revolutionary progress in Automatic Speech Recognition (ASR). However, they have not been totally proven to produce better performance on tasks other than ASR. In this work, we explored partial fine-tuning and entire fine-tuning on wav2vec 2.0 and HuBERT pre-trained models for three non-ASR speech tasks: Speech Emotion Recognition, Speaker Verification and Spoken Language Understanding. With simple proposed downstream frameworks, the best scores reached 79.58\% weighted accuracy on speaker-dependent setting and 73.01\% weighted accuracy on speaker-independent setting for Speech Emotion Recognition on IEMOCAP, 2.36\% equal error rate for Speaker Verification on VoxCeleb1, 89.38\% accuracy for Intent Classification and 78.92\% F1 for Slot Filling on SLURP, showing the strength of fine-tuned wav2vec 2.0 and HuBERT on learning prosodic, voice-print and semantic representations.
\end{abstract}
\section{Introduction}

Nowadays, people are expecting less labeled data to train well-generalized models for supervised tasks, since data labeling is a very time- and money-consuming process. Furthermore, people have been attempting to find a powerful feature embedding that can assist the fine-tuning and multi-task training for downstream tasks. The appearance of self-supervised learning meets the above two requirements exactly. In speech domain, excellent self-supervised models are emerging \cite{APC, NPC, MOCKINGJAY, TERA, PASE+, wav2vec, vq-wav2vec, wavLM, unispeech}, among which the most high-performing and the most widely used are wav2vec 2.0 \cite{wav2vec2} and HuBERT \cite{HuBERT}. Many wav2vec 2.0/HuBERT pretrained models have also been published and this has greatly promoted their applications in the field of speech. Therefore, we chose wav2vec2.0 and HuBERT as our research objects in this work.

\begin{figure*}[t]
  \centering
  \includegraphics[width=\linewidth]{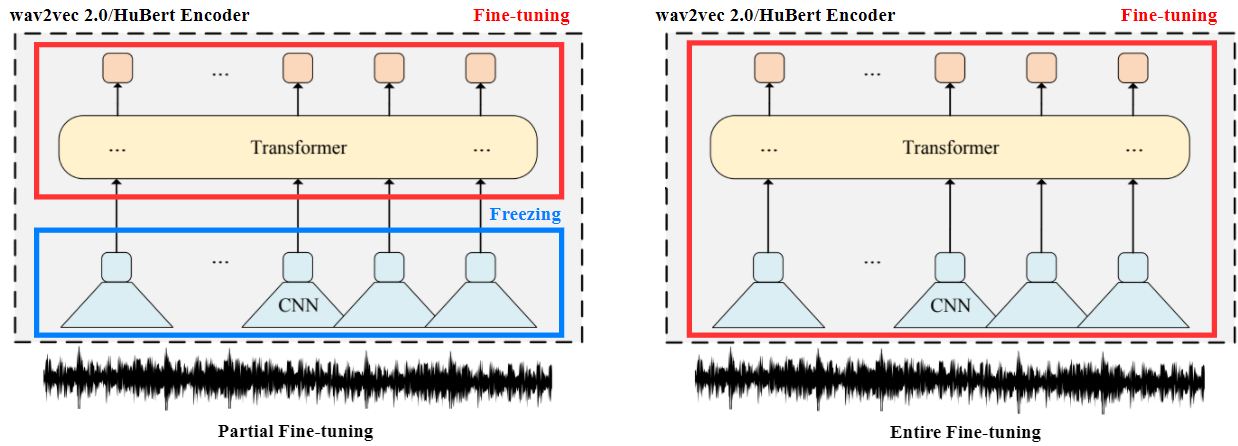}
  \caption{Partial fine-tuning (left) and entire fine-tuning (right) of wav2vec 2.0/HuBERT.}
  \label{fig:speech_production}
\end{figure*}

The wav2vec 2.0 model architecture contains mainly three modules. A convolutional neural network (CNN) \textbf{feature encoder} encodes the raw waveform inputs into latent speech representations. Mask operations are applied before they are fed to the Transformer-based \textbf{contextualized encoder}. A \textbf{quantization module} is used to quantize the latent speech representations from the CNN encoder into a discretized embedding which is then used as the target. The objective is to optimize the contrastive loss, which enforces the model to identify the true quantized latent speech representations.

HuBERT shares the same architecture as wav2vec 2.0. Instead of constructing a contrastive loss, HuBERT uses an offline clustering step to generate noisy labels for Masked Language Model pretraining. Specifically, HuBERT consumes masked continuous speech features to predict predetermined cluster assignments. The predictive loss is applied over the masked regions, forcing the model to learn good high-level representations of unmasked inputs in order to infer the targets of masked ones correctly.

Wav2vec 2.0 and HuBERT outperformed all existing ASR models at that time, proving that they can construct a better verbal embedding. However, speech also contains other important information such as emotion, speaker and semantics, for which the industry also has high expectations. In the field of \textbf{Speech Emotion Recognition} (\textbf{SER}), \textbf{Speaker Verification} (\textbf{SV}) and \textbf{Spoken Language Understanding} (\textbf{SLU}), it is still vague whether self-supervised models can produce better performance compared with traditional supervised models (spectral features + CNN-based feature extraction + RNN/Transformer based time series modeling) \cite{tr-emo-1, tr-emo-2, tr-sv-1, tr-sv-2, tr-slu-1}. However, meaningful attempts have been made in some previous works, which we will introduce below.

In SUPERB \cite{superb}, the performance of different frozen self-supervised encoders were benchmarked across a wide range of speech tasks. For SER, the HuBERT large model stood out from other self-supervised encoders with 67.62\% accuracy (ACC) on IEMOCAP \cite{IEMOCAP}. For SV, the HuBERT base model obtained the best Equal Error Rate (EER) 5.11\% on VoxCeleb1 \cite{VoxCeleb}. SLU contains two separate subtasks: \textbf{Intent Classification} (\textbf{IC}) and \textbf{Slot Filling}   (\textbf{SF}). The HuBERT large model achieved the best results on both IC and SF tasks with 98.76\% ACC on Fluent Speech Commands dataset \cite{FSC} and 89.81\% F1 score on SNIPS \cite{SNIPS} respectively.

For SER, \cite{Interspeech-emotion} combined the features from frozen wav2vec2.0 with other hand-crafted prosodic features and then fed them into a 1d-CNN for a deeper extraction. \cite{Interspeech-emotion-2} explored wav2vec fine-tuning strategies and 65.4\% WA on IEMOCAP was achieved. For SV, \cite{Interspeech-speaker, speaker-sota} both explored fine-tuned wav2vec 2.0, \cite{Interspeech-speaker} obtained 3.61\% EER on VoxCeleb1, while \cite{speaker-sota} also obtained 1.91\% EER on VoxCeleb1 by adding VoxCeleb2 into the training set.


We notice that self-supervised models were only used as frozen feature extractors in SUPERB and some other works. Believing that only by fine-tuning can we show the real power of self-supervised models, we explored the fine-tuning of wav2vec2.0/HuBERT on three speech tasks and provided full fine-tuning experiment details. Taking inspiration from \cite{wav2vec2} and \cite{HuBERT}, we added another fine-tuning method by splitting a pre-trained wav2vec 2.0/HuBERT model into two parts: the CNN feature encoder and the Transformer contextualized encoder. We froze the CNN feature encoder and only fine-tuned the Transformer contextualized encoder. We then tested partially fine-tuned wav2vec2.0/HuBERT pre-trained models together with the entirely fine-tuned ones with the following tasks below: 
\begin{itemize}
    \item  Speech Emotion Recognition on IEMOCAP
    \item  Speaker Verification on VoxCeleb1
    \item  Spoken Language Understanding on SLURP \cite{SLURP}
\end{itemize}

The results show that our fine-tuned models achieved excellent results on the three tasks, which further proves their strong capacity on constructing problem-agnostic representations. The code and fine-tuned models for SER and SLU have been open-sourced on SpeechBrain \cite{SpeechBrain} \footnote{https://github.com/speechbrain/speechbrain/tree/develop/recipes}.

\section{Method}

In this section, we will first introduce the pre-training of wav2vec 2.0/HuBERT model, then we will show our fine-tuning methods and downstream models for each task.

 \subsection{Pretrained wav2vec 2.0}
The wav2vec 2.0 pre-training is similar to the masked language modelling in BERT \cite{Bert} and is carried out under a self-supervised setting. Contiguous time steps from the CNN encoder representations are randomly masked, and the model is trained to reproduce the quantized local encoder representations for masked frames at the output of the contextualized encoder.

\begin{equation}
  L_m = -\log \frac{\exp(sim(c_t, q_t)/ \kappa)}
  {\sum_{\tilde{q} \in Q_t}(\exp(sim(c_t, \tilde{q} )/\kappa)}
  \label{eq3}
\end{equation}

The training objective is illustrated in Eq.1, where $sim(c_t, q_t)$ is the cosine similarity between the contextualized encoder outputs $c_t$ and the quantized CNN encoder representations $q_t$, $t$ is the masked time step, $Q_t$ is the union of candidate representations $\tilde{q}$ which includes $q_t$ and $K = 100$ distractors, $\kappa$ is the temperature which is set to 0.1. The distractors are outputs of the local encoder sampled from masked frames belonging to the same utterance as $q_t$. The contrastive loss is then given by $L_m$ summed over all masked frames. At the end, an $L2$ regularization is added to the contrastive loss, as well as a diversity loss to increase the use of the quantized codebook representations.

The pre-training process is optimized with Adam \cite{Adam} and the learning rate decays linearly after a waming up. In \cite{wav2vec2}, wav2vec 2.0 is also fine-tuned on ASR aiming to improve ASR performance. For ASR fine-tuning, a randomly initialized linear projection is added to the output of the contextual encoder and the CTC (Connectionist Temporal Classification \cite{CTC}) loss is minimized. For more details of the pre-training and ASR fine-tuning of wav2vec 2.0, please refer to \cite{wav2vec2}.

In this work, we compare four released wav2vec 2.0 pre-trained models: the wav2vec 2.0 base model (12 transformer blocks and 768 embedding dimension) and its ASR fine-tuned version, the wav2vec 2.0 large model (24 transformer blocks and 1024 embedding dimension) and its ASR fine-tuned version. Both base and large models are pre-trained on 960h LibriSpeech \cite{LibriSpeech} data, which is also used for their ASR fine-tuning. ASR fine-tuned models for both wav2vec 2.0 and HuBERT are taken into consideration because we assume that some tasks may benefit from the ASR fine-tuning.

\subsection{Pretrained HuBERT}
In the same way as wav2vec 2.0, CNN-encoded audio features are randomly masked in HuBERT. To generate labels for the first iteration of HuBERT pre-training, a k-means clustering is applied on 39-dimensional MFCC features. To generate better targets for the subsequent iterations, k-means clustering then works on the latent features extracted from the HuBERT model pre-trained in the previous iteration. A projection layer is added over transformer blocks to predict cluster labels. Cross-entropy loss is computed over masked timestamps, which can be defined as: 
\begin{equation}
  L_m(f;X, \{Z^{(k)}\}_k, M)= \sum_{t \in M}\sum_{k} \log p_f^{(k)}(z_t^{(k)} | \widetilde{X}, t)
\end{equation}
$ M \subset [T]$ denotes the set of indices to be masked for a length-$T$ sequence $X$, and $\widetilde{X}= r(X;M)$ denotes a corrupted version of $X$ where $x_t$ is replaced with a mask embedding $\widetilde{x}$ if $t \in M$. A masked prediction model $f$ takes as input $\widetilde{X}$ and predicts a distribution over the target indices at each timestep $p_f (\cdot |\widetilde{X}; t)$. To improve target quality, cluster ensembles are utillized in case that an individual clustering model performs badly, $Z^{(k)}$ then denotes the target sequences generated by the $k$-th clustering model.

HuBERT pre-training uses the same optimizer and learning rate scheduler as wav2vec 2.0. For ASR fine-tuning, the projection layer is removed and replaced by a randomly initialized softmax layer, then the CTC loss is optimized. For more details of the pre-training of HuBERT, please refer to \cite{HuBERT}.

Like wav2vec 2.0, we compare three released HuBERT pre-trained models: the HuBERT base model (12 transformer blocks and 768 embedding dimension, of which no ASR fine-tuned version is released), the HuBERT large model (24 transformer blocks and 1024 embedding dimension) and its ASR fine-tuned version. The HuBERT base model is pre-trained on 960h LibriSpeech data, while the large model is pre-trained on 60k hours Libri-Light \cite{LibriLight} data. The ASR fine-tuning is also based on 960h LibriSpeech data.

\begin{figure}[t]
  \centering
  \includegraphics[width=\linewidth]{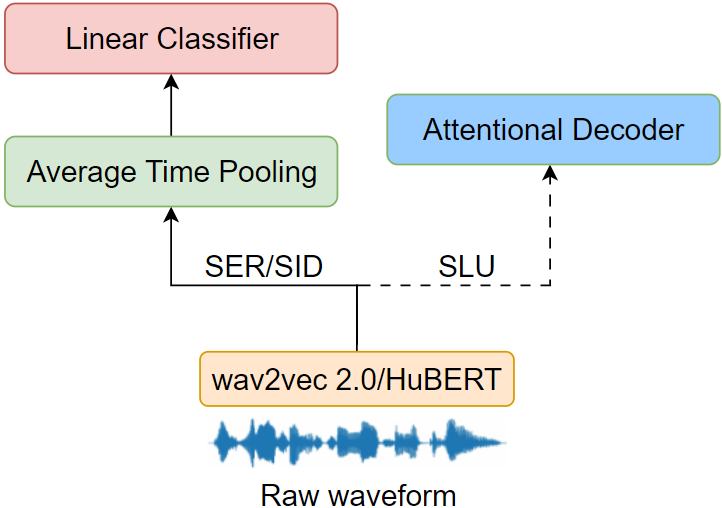}
  \caption{Simple downstream models for SER, SID and SLU. For SER and SID, an average time pooling and a linear classifier is built over wav2vec 2.0/HuBERT. For SLU, an attentional decoder decodes intents and slots directly from the fine-tuned wav2vec2.0/HuBERT embedding.}
  \label{fig:speech_production}
\end{figure}

\subsection{Fine-tuning}
As is shown in Figure 1 on the left for partial fine-tuning, the wav2vec 2.0/HuBERT model is divided into two parts: the CNN-based feature encoder and the transformer-based contextualized encoder. We froze the CNN-based feature encoder, fixing all the parameters of these CNN blocks, and only fine-tuned the parameters of the transformer blocks. Partial fine-tuning can be understood as a domain adaptation training for the top level, which aims to prevent interference and damage to the bottom CNN layers that already have an expressive ability.

For entire fine-tuning which is shown on the right in Figure 1, the CNN and Transformer modules are both fine-tuned during the downstream training process. By training general features at the bottom level, entire fine-tuning allows higher-level expressions to be more complete and more targeted.


Then, assuming that fine-tuned wav2vec 2.0/HuBERT are already powerful enough to capture information, we directly added simple downstream adaptors (classifier/decoder) to wav2vec 2.0/HuBERT without adding another heavy and redundant encoder. The downstream adaptors for each task are presented as below.

For SER, an average time pooling and one linear layer are added as a simple downstream classifier (Fig.2). The average time pooling compresses variant time lengths into one, then the linear layer effectuates an utterance-level classification minimizing the cross-entropy loss.

For SV, a \textbf{Speaker Identification} (\textbf{SID}) task is first implemented using the same downstream framework as SER. Pairwise cosine-similarity scores are then produced for SV on the pre-trained SID embeddings before the linear classification layer.

For SLU (Fig.2), another attentional GRU-based decoder is added to decode semantic information directly from the fine-tuned wav2vec2.0/HuBERT embedding. In our work, intents and slots are both treated as a sequence-to-sequence ASR task and are both decoded from the attentional decoder. The Negative Log-Likelihood (NLL) loss is then calculated over a character-level token generation. Following the observations of \cite{timers_and_such}, we utilized a beam-search with a beam of 80 without coverage penalty to identify the optimum sequence for validation set and test set.

\section{Experiments}

In the experiment section, we will first introduce the datasets used for the three tasks, then we will list the models we compared and add details of our experiment settings. Finally, we will show the results for each task and compare with the existing state-of-the-art baselines.

\subsection{Datasets}
The three most widely used and most representative datasets were chosen in our experiments, which are IEMOCAP for SER, VoxCeleb1 for SV and SLURP for SLU.

The Interactive Emotional Dyadic Motion Capture (IEMOCAP) dataset has approximately 12 hours of data and consists of scripted and improvised dialogues by 10 speakers. In order to form a contrast in this work, we used 4 emotional classes as in SUPERB: anger, happiness, sadness and neutral, following the work of \cite{IEMOCAP-relabel}. The evaluation metric is weighted accuracy (WA) and the experiments were carried out on two different split settings: Speaker-Dependent (\textbf{SD}) setting and Speaker-Independent (\textbf{SI}) setting. For SD, the results were averaged on 5 different random seeds for train-validation-test split. For SI, a 10-fold cross-validation was performed with a leave-two-speaker-out strategy (one for validation and one for test).



VoxCeleb1 contains over 100,000 utterances from 1,251 speakers, with approximately 351 hours of audio in total. In our work, a Speaker Identification task was first implemented, encouraging the model to learn to distinguish 1211 different voice-prints. A verification was then carried out on the vox1-o test set of 40 speakers by calculating cosine similarity on the embeddings from the pre-trained Speaker Identification model. VoxCeleb2 and noise augmentation were not used in our experiments. We used equal error rate (EER) as the evaluation metric and the results were averaged on 5 different seeds for train-validation split.


The Spoken Language Understanding Resource Package (SLURP) Dataset is a collection of 72K audio recordings of single turn user interactions with a home assistant, annotated with three levels of semantics: Scenario, Action and Entities. The training and evaluation are based on its official training, validation and test sets.

\subsection{Fine-tuning settings}
We rename the models we compare with a method as below.
\begin{itemize}
    \item  EF/PF/Frozen: Entirely Fine-tuned/Partially Fine-tuned/Not fine-tuned
    \item  w2v/hbt: wav2vec 2.0/HuBERT based model
    \item  base/large: base/large pre-trained model
    \item  -/960h: with/without ASR fine-tuning using 960h LibriSpeech data
\end{itemize}

For example, \textit{EF-w2v-base} refers to an entirely fine-tuned wav2vec 2.0 base model, while \textit{PF-hbt-large-960h} refers to a partially fine-tuned HuBERT large model with an ASR fine-tuning. For more detailed parameters of released pre-trained wav2vec 2.0/HuBERT models, please refer to \cite{wav2vec2} and \cite{HuBERT}.

During the fine-tuning process, we applied two different schedulers to respectively adjust the fine-tuning learning rate of the wav2vec 2.0/HuBERT encoder and the learning rate of the downstream model. Both the schedulers use an Adam Optimizer and linearly anneal the learning rates according to the performance of validation stage. For SER and SV, the initialized fine-tuning learning rate and the downstream learning rate are set to $10^{-5}$ and $10^{-4}$. For SLU, these values are set to $10^{-5}$ and $3 \times 10^{-4}$.

\begin{table}[th]
  \centering
  \caption{Benchmark results for two utterance-level tasks: Speech Emotion Recognition (SER) on weighted accuracy (WA\%) and Speaker Verification (SV) on Equal Error Rate (EER\%). In relation to SER, it is divided into SER-SD (Speaker-Dependent settng) and SER-SI (Speaker-Independent setting).
}
  \label{tab:example}
  \centering
  \begin{tabular}{cccc}
    \toprule
    \multicolumn{1}{c}{\textbf{Model}} &   \multicolumn{1}{c}{\textbf{SER-SD}}   &
    \multicolumn{1}{c}{\textbf{SER-SI}}   &
    \multicolumn{1}{c}{\textbf{SV}}\\
    \midrule
    EF-w2v-base          & 75.90 & 70.75 & 2.77 \\
    PF-w2v-base          & 77.02 & 70.21 & 3.15 \\
    EF-w2v-base-960h     & 73.64 & 64.20 & 4.46  \\
    PF-w2v-base-960h     & 73.84 & 68.34 & 4.38  \\
    \midrule
    EF-w2v-large          & 77.00 & 70.96 & 3.42  \\
    PF-w2v-large          & 77.47 & 70.99 & 3.85  \\
    EF-w2v-large-960h     & 73.00 & 68.18 & 4.27  \\
    PF-w2v-large-960h     & 76.75 & 69.08 & 4.47  \\
    \midrule
    EF-hbt-base          & 76.53 & 69.83 & 2.84 \\
    PF-hbt-base          & 76.60 & 69.68 & 3.13 \\
    \midrule
    EF-hbt-large          & 78.52 & 72.31 & 2.86  \\
    PF-hbt-large          & \textbf{79.58} & \textbf{73.01} & 3.21  \\
    EF-hbt-large-960h     & 78.78 & 72.71 & \textbf{2.36} \\
    PF-hbt-large-960h     & 78.96 & 72.98 & 2.38 \\
    \midrule
    Frozen-w2v-base\cite{superb} & - & 63.43 & 6.02 \\
    Frozen-w2v-large\cite{superb} & - & 65.64 & 5.65 \\
    Frozen-hbt-base\cite{superb} & - & 64.92 & 5.11 \\
    Frozen-hbt-large\cite{superb} & - & 67.62 & 5.98 \\
    \midrule
    Head Fusion\cite{head-fusion} & 76.18 & - & -  \\
    Attention Pooling\cite{attention_pooling} & - & 71.75 & -  \\
    Siamese Capsule\cite{siamese} & - & - & 3.14  \\
    \bottomrule
  \end{tabular}
  
\end{table}

\subsection{Results and discussion}
\subsubsection{Speech Emotion Recognition \& Speaker Verification}
The results of SER and SV are shown in Table 1. For comparison, we show SUPERB's results as a non-fine-tuned baseline (marked with \cite{superb} in Table 1). Furthermore, we took Head-Fusion ACNN \cite{head-fusion} for SER-SD (Speaker-Dependent setting), Attention Pooling based representation \cite{attention_pooling} for SER-SI (Speaker-Independent setting) and Siamese Capsule network \cite{siamese} for SV as state-of-the-art baselines respectively. Compared with other more recent works, \cite{attention_pooling} provides a comparable result by reporting a competitve Weighted Accuracy using only speech, and \cite{siamese} also provides a comparable result using only Voxceleb1 as the training set (the same as SUPERB). First of all, from an overall perspective, we notice a significant improvement on the results of fine-tuned models over those of SUPERB's non-fine-tuned models. Then, for SER, we are surprised to find that all of our fine-tuned models performed well, where the partially fine-tuned HuBERT large model reached a best WA of 79.58\% for SER-SD and a best WA of 73.01\% for SER-SI, improving by 3.40\% and 1.26\% on the state-of-the-art baselines respectively. Moreover, we observe that partial fine-tuning appeared to be a better fine-tuning method than entire fine-tuning. We consider that IEMOCAP is a small dataset with only 12 hours of data and training too many parameters may easily cause an overfitting. Additionally, we noticed that the ASR fine-tuning was not helping the downstream SER task, suggesting a loss of prosodic information during the ASR fine-tuning. 

In the case of SV, the entirely fine-tuned HuBERT with ASR fine-tuning reached a best 2.36\% Equal Error Rate, surpassing the baseline by 0.78\%. However, contrary to SER, entire fine-tuning outperforms partial fine-tuning as can be seen from the results. Due to the large amount of data (351 hours) of VoxCeleb1 that are also acoustically similar to the data used for pre-training, the pre-trained encoder parameters provide an ideal initialization for the downstream SV task, releasing all the layers and fine-tuning with a low learning rate can lead to a good result. Finally, we find that HuBERT turned out to be a better self-supervised encoder compared to wav2vec 2.0 for both SER and SV tasks.

\begin{table}[th]
  \centering
  \caption{Benchmark results for Spoken Language Understanding: Intent Classification (IC) on accuracy (ACC\%) and Slot Filling (SF) on F1 score (F1\%).
}
  \label{tab:example}
  \centering
  \begin{tabular}{ccc}
    \toprule
    \multicolumn{1}{c}{\textbf{Model}} &   \multicolumn{1}{c}{\textbf{IC (ACC\%)}}   &
    \multicolumn{1}{c}{\textbf{SF (F1\%)}}\\
    \midrule
    EF-w2v-base          & 87.13 & 74.32 \\
    PF-w2v-base          & 86.58 & 74.73 \\
    EF-w2v-base-960h     & 85.89 & 74.33  \\
    PF-w2v-base-960h     & 86.13 & 73.78  \\
    \midrule
    EF-w2v-large          & 85.80 & 72.45 \\
    PF-w2v-large          & 86.29 & 73.16 \\
    EF-w2v-large-960h     & 86.10 & 73.39  \\
    PF-w2v-large-960h     & 86.35 & 74.03  \\
    \midrule
    EF-hbt-base          & 87.44 & 75.06 \\
    PF-hbt-base          & 87.51 & 75.32 \\
    \midrule
    EF-hbt-large          &  \textbf{89.38} & 78.43 \\
    PF-hbt-large          & 89.22 & \textbf{78.92} \\
    EF-hbt-large-960h     & 88.71 & 78.89  \\
    PF-hbt-large-960h     & 88.32 & 78.17  \\
    \midrule
    Frozen-w2v-base & 47.15 & 37.66 \\
    Frozen-w2v-large & 3.88 & 3.85 \\
    Frozen-hbt-base & 68.74 & 57.06 \\
    Frozen-hbt-large & 74.42 & 60.07 \\
    
    \midrule
    CTI\cite{CTI} & 86.92 & 74.66  \\
    \bottomrule
  \end{tabular}
  
\end{table}

\subsubsection{Spoken Language Understanding}
For SLU, the results of its two subtasks are shown in Table 2. Likewise, we carried out experiments of frozen wav2vec 2.0/HuBERT models to form a contrast. However, the performance of frozen models on this task drops significantly, especially for wav2vec 2.0 large model the loss cannot even converge, demonstrating that the frozen wav2vec 2.0/HuBERT cannot hold complete semantic information. Continuous Token Interface \cite{CTI} is chosen as the state-of-the-art baseline. The best ACC for IC is 89.38\% with the entirely fine-tuned HuBERT large model, while the partially fine-tuned HuBERT large model reached the best F1 78.92\% for SF. For SLU, the gap between the two fine-tuning methods is not obvious. A slight drop is observed on ASR fine-tuned models, which implies that ASR fine-tuning will also result in a loss of semantic information.
  

\section{Conclusions}
\label{sec:conclusions}

In this work we explored different fine-tuning methods on two of the most powerful self-supervised models (wav2vec 2.0 and HuBERT), then benchmarked their performance on Speech Emotion Recognition, Speaker Verification and Spoken Language Understanding tasks. State-of-the-art results were achieved for all the three tasks, proving the excellent generalizability of wav2vec 2.0/HuBERT on learning prosodic, voice-print and semantic representations. We hope to show the broad prospects of self-supervised learning and also provide some useful insights for its industrial applications. 


\clearpage

\bibliographystyle{IEEEbib}
\bibliography{mybib}

\begin{thebibliography}{10}

\bibitem{APC}
Yu-An Chung, Wei-Ning Hsu, Hao Tang, and James Glass,
\newblock ``An unsupervised autoregressive model for speech representation
  learning,''
\newblock in {\em Interspeech}, 2019.

\bibitem{NPC}
Alexander~H. Liu, Yu-An Chung, and James Glass,
\newblock ``{Non-Autoregressive Predictive Coding for Learning Speech
  Representations from Local Dependencies},''
\newblock in {\em Proc. Interspeech 2021}, 2021, pp. 3730--3734.

\bibitem{MOCKINGJAY}
Andy~T Liu, Shu-wen Yang, Po-Han Chi, Po-chun Hsu, and Hung-yi Lee,
\newblock ``Mockingjay: Unsupervised speech representation learning with deep
  bidirectional transformer encoders,''
\newblock in {\em ICASSP 2020-2020 IEEE International Conference on Acoustics,
  Speech and Signal Processing (ICASSP)}. IEEE, 2020, pp. 6419--6423.

\bibitem{TERA}
Andy~T Liu, Shang-Wen Li, and Hung-yi Lee,
\newblock ``Tera: Self-supervised learning of transformer encoder
  representation for speech,''
\newblock {\em IEEE/ACM Transactions on Audio, Speech, and Language
  Processing}, vol. 29, pp. 2351--2366, 2021.

\bibitem{PASE+}
Mirco Ravanelli, Jianyuan Zhong, Santiago Pascual, Pawel Swietojanski, Joao
  Monteiro, Jan Trmal, and Yoshua Bengio,
\newblock ``Multi-task self-supervised learning for robust speech
  recognition,''
\newblock in {\em ICASSP 2020-2020 IEEE International Conference on Acoustics,
  Speech and Signal Processing (ICASSP)}. IEEE, 2020, pp. 6989--6993.

\bibitem{wav2vec}
Steffen Schneider, Alexei Baevski, Ronan Collobert, and Michael Auli,
\newblock ``{wav2vec: Unsupervised Pre-Training for Speech Recognition},''
\newblock in {\em Proc. Interspeech 2019}, 2019, pp. 3465--3469.

\bibitem{vq-wav2vec}
A.~Baevski, S.~Schneider, and M.~Auli,
\newblock ``vq-wav2vec: Self-supervised learning of discrete speech
  representations,''
\newblock in {\em International Conference on Learning Representations (ICLR)},
  2020.

\bibitem{wavLM}
Sanyuan Chen, Chengyi Wang, Zhengyang Chen, Yu~Wu, Shujie Liu, Zhuo Chen, Jinyu
  Li, Naoyuki Kanda, Takuya Yoshioka, Xiong Xiao, et~al.,
\newblock ``Wavlm: Large-scale self-supervised pre-training for full stack
  speech processing,''
\newblock {\em IEEE Journal of Selected Topics in Signal Processing}, 2022.

\bibitem{unispeech}
Chengyi Wang, Yu~Wu, Yao Qian, Kenichi Kumatani, Shujie Liu, Furu Wei, Michael
  Zeng, and Xuedong Huang,
\newblock ``Unispeech: Unified speech representation learning with labeled and
  unlabeled data,''
\newblock in {\em International Conference on Machine Learning}. PMLR, 2021,
  pp. 10937--10947.

\bibitem{wav2vec2}
Alexei Baevski, Yuhao Zhou, Abdelrahman Mohamed, and Michael Auli,
\newblock ``wav2vec 2.0: {A} framework for self-supervised learning of speech
  representations,''
\newblock in {\em NeurIPS}, 2020.

\bibitem{HuBERT}
Wei-Ning Hsu, Benjamin Bolte, Yao-Hung~Hubert Tsai, Kushal Lakhotia, Ruslan
  Salakhutdinov, and Abdelrahman Mohamed,
\newblock ``Hubert: Self-supervised speech representation learning by masked
  prediction of hidden units,''
\newblock {\em arXiv preprint arXiv:2106.07447}, 2021.

\bibitem{tr-emo-1}
Pengcheng Li, Yan Song, Ian~Vince McLoughlin, Wu~Guo, and Lirong Dai,
\newblock ``An attention pooling based representation learning method for
  speech emotion recognition,''
\newblock in {\em {Interspeech}}. 2018, pp. 3087--3091, {ISCA}.

\bibitem{tr-emo-2}
Xixin Wu, Songxiang Liu, Yuewen Cao, Xu~Li, Jianwei Yu, Dongyang Dai, Xi~Ma,
  Shoukang Hu, Zhiyong Wu, Xunying Liu, et~al.,
\newblock ``Speech emotion recognition using capsule networks,''
\newblock in {\em ICASSP 2019-2019 IEEE International Conference on Acoustics,
  Speech and Signal Processing (ICASSP)}. IEEE, 2019, pp. 6695--6699.

\bibitem{tr-sv-1}
Gautam Bhattacharya, Md~Jahangir Alam, and Patrick Kenny,
\newblock ``Deep speaker embeddings for short-duration speaker verification.,''
\newblock in {\em Interspeech}, 2017, pp. 1517--1521.

\bibitem{tr-sv-2}
Brecht Desplanques, Jenthe Thienpondt, and Kris Demuynck,
\newblock ``{ECAPA-TDNN:} emphasized channel attention, propagation and
  aggregation in {TDNN} based speaker verification,''
\newblock in {\em Interspeech}. 2020, pp. 3830--3834, {ISCA}.

\bibitem{tr-slu-1}
Dmitriy Serdyuk, Yongqiang Wang, Christian Fuegen, Anuj Kumar, Baiyang Liu, and
  Yoshua Bengio,
\newblock ``Towards end-to-end spoken language understanding,''
\newblock in {\em 2018 IEEE International Conference on Acoustics, Speech and
  Signal Processing (ICASSP)}. IEEE, 2018, pp. 5754--5758.

\bibitem{superb}
Shu-wen Yang, Po-Han Chi, Yung-Sung Chuang, Cheng-I Lai, Kushal Lakhotia, Yist
  Lin, Andy Liu, Jiatong Shi, Xuankai Chang, Guan-Ting Lin, Tzu-Hsien Huang,
  Wei-Cheng Tseng, Ko-tik Lee, Da-Rong Liu, Zili Huang, Shuyan Dong, Shang-Wen
  Li, Shinji Watanabe, Abdelrahman Mohamed, and Hung-yi Lee,
\newblock ``Superb: Speech processing universal performance benchmark,''
\newblock in {\em Proc. Interspeech 2021}, 08 2021, pp. 1194--1198.

\bibitem{IEMOCAP}
Carlos Busso, Murtaza Bulut, Chi-Chun Lee, Abe Kazemzadeh, Emily Mower, Samuel
  Kim, Jeannette~N Chang, Sungbok Lee, and Shrikanth~S Narayanan,
\newblock ``Iemocap: Interactive emotional dyadic motion capture database,''
\newblock {\em Language resources and evaluation}, vol. 42, no. 4, pp.
  335--359, 2008.

\bibitem{VoxCeleb}
Arsha Nagrani, Joon~Son Chung, and Andrew Zisserman,
\newblock ``Voxceleb: {A} large-scale speaker identification dataset,''
\newblock in {\em Interspeech}. 2017, pp. 2616--2620, {ISCA}.

\bibitem{FSC}
Natalia Tomashenko, Antoine Caubri{\`e}re, Yannick Est{\`e}ve, Antoine Laurent,
  and Emmanuel Morin,
\newblock ``Recent advances in end-to-end spoken language understanding,''
\newblock in {\em International Conference on Statistical Language and Speech
  Processing}. Springer, 2019, pp. 44--55.

\bibitem{SNIPS}
Alice Coucke, Alaa Saade, Adrien Ball, Th{\'e}odore Bluche, Alexandre Caulier,
  David Leroy, Cl{\'e}ment Doumouro, Thibault Gisselbrecht, Francesco
  Caltagirone, Thibaut Lavril, et~al.,
\newblock ``Snips voice platform: an embedded spoken language understanding
  system for private-by-design voice interfaces,''
\newblock {\em arXiv preprint arXiv:1805.10190}, 2018.

\bibitem{Interspeech-emotion}
Leonardo Pepino, Pablo Riera, and Luciana Ferrer,
\newblock ``{Emotion Recognition from Speech Using wav2vec 2.0 Embeddings},''
\newblock in {\em Proc. Interspeech 2021}, 2021, pp. 3400--3404.

\bibitem{Interspeech-emotion-2}
Yangyang Xia, Li-Wei Chen, Alexander Rudnicky, and Richard~M Stern,
\newblock ``Temporal context in speech emotion recognition,''
\newblock in {\em Proc. Interspeech}, 2021, vol. 2021, pp. 3370--3374.

\bibitem{Interspeech-speaker}
Zhiyun Fan, Meng Li, Shiyu Zhou, and Bo~Xu,
\newblock ``{Exploring wav2vec 2.0 on Speaker Verification and Language
  Identification},''
\newblock in {\em Proc. Interspeech 2021}, 2021, pp. 1509--1513.

\bibitem{speaker-sota}
Nik Vaessen and David~A Van~Leeuwen,
\newblock ``Fine-tuning wav2vec2 for speaker recognition,''
\newblock in {\em ICASSP 2022-2022 IEEE International Conference on Acoustics,
  Speech and Signal Processing (ICASSP)}. IEEE, 2022, pp. 7967--7971.

\bibitem{SLURP}
Emanuele Bastianelli, Andrea Vanzo, Pawel Swietojanski, and Verena Rieser,
\newblock ``Slurp: A spoken language understanding resource package,''
\newblock {\em arXiv preprint arXiv:2011.13205}, 2020.

\bibitem{SpeechBrain}
Mirco Ravanelli, Titouan Parcollet, Peter Plantinga, Aku Rouhe, Samuele
  Cornell, Loren Lugosch, Cem Subakan, Nauman Dawalatabad, Abdelwahab Heba,
  Jianyuan Zhong, et~al.,
\newblock ``Speechbrain: A general-purpose speech toolkit,''
\newblock {\em arXiv preprint arXiv:2106.04624}, 2021.

\bibitem{Bert}
Jacob Devlin, Ming{-}Wei Chang, Kenton Lee, and Kristina Toutanova,
\newblock ``{BERT:} pre-training of deep bidirectional transformers for
  language understanding,''
\newblock in {\em {NAACL-HLT} {(1)}}. 2019, pp. 4171--4186, Association for
  Computational Linguistics.

\bibitem{Adam}
Diederik~P. Kingma and Jimmy Ba,
\newblock ``Adam: {A} method for stochastic optimization,''
\newblock in {\em {ICLR} (Poster)}, 2015.

\bibitem{CTC}
Alex Graves, Santiago Fern{\'a}ndez, Faustino Gomez, and J{\"u}rgen
  Schmidhuber,
\newblock ``Connectionist temporal classification: labelling unsegmented
  sequence data with recurrent neural networks,''
\newblock in {\em Proceedings of the 23rd international conference on Machine
  learning}, 2006, pp. 369--376.

\bibitem{LibriSpeech}
Vassil Panayotov, Guoguo Chen, Daniel Povey, and Sanjeev Khudanpur,
\newblock ``Librispeech: An asr corpus based on public domain audio books,''
\newblock in {\em 2015 IEEE International Conference on Acoustics, Speech and
  Signal Processing (ICASSP)}, 2015, pp. 5206--5210.

\bibitem{LibriLight}
Jacob Kahn, Morgane Rivi{\`e}re, Weiyi Zheng, Evgeny Kharitonov, Qiantong Xu,
  Pierre-Emmanuel Mazar{\'e}, Julien Karadayi, Vitaliy Liptchinsky, Ronan
  Collobert, Christian Fuegen, et~al.,
\newblock ``Libri-light: A benchmark for asr with limited or no supervision,''
\newblock in {\em ICASSP 2020-2020 IEEE International Conference on Acoustics,
  Speech and Signal Processing (ICASSP)}. IEEE, 2020, pp. 7669--7673.

\bibitem{timers_and_such}
Loren Lugosch, Piyush Papreja, Mirco Ravanelli, Abdelwahab Heba, and Titouan
  Parcollet,
\newblock ``{Timers and Such: A Practical Benchmark for Spoken Language
  Understanding with Numbers},''
\newblock {\em NeurIPS Datasets and Benchmarks}, 2021.

\bibitem{IEMOCAP-relabel}
Haytham~M Fayek, Margaret Lech, and Lawrence Cavedon,
\newblock ``Evaluating deep learning architectures for speech emotion
  recognition,''
\newblock {\em Neural Networks}, vol. 92, pp. 60--68, 2017.

\bibitem{head-fusion}
Mingke Xu, Fan Zhang, and Wei Zhang,
\newblock ``Head fusion: Improving the accuracy and robustness of speech
  emotion recognition on the iemocap and ravdess dataset,''
\newblock {\em IEEE Access}, vol. 9, pp. 74539--74549, 2021.

\bibitem{attention_pooling}
Pengcheng Li, Yan Song, Ian~Vince McLoughlin, Wu~Guo, and Li-Rong Dai,
\newblock ``An attention pooling based representation learning method for
  speech emotion recognition,''
\newblock {\em Interspeech}, 2018.

\bibitem{siamese}
Amirhossein Hajavi and Ali Etemad,
\newblock ``Siamese capsule network for end-to-end speaker recognition in the
  wild,''
\newblock in {\em ICASSP 2021-2021 IEEE International Conference on Acoustics,
  Speech and Signal Processing (ICASSP)}. IEEE, 2021, pp. 7203--7207.

\bibitem{CTI}
Seunghyun Seo, Donghyun Kwak, and Bowon Lee,
\newblock ``Integration of pre-trained networks with continuous token interface
  for end-to-end spoken language understanding,''
\newblock {\em arXiv preprint arXiv:2104.07253}, 2021.

\end{thebibliography}

\end{document}